\documentclass[10pt,twocolumn,letterpaper]{article}

\usepackage{wacv}
\usepackage{times}
\usepackage{epsfig}
\usepackage{graphicx}
\usepackage{amsmath}
\usepackage{amssymb}
\usepackage[accsupp]{axessibility}  
\usepackage{subcaption}
\usepackage[ruled]{algorithm2e}
\usepackage{multirow}
\usepackage{subcaption}
\usepackage{kotex}

\usepackage{tabularx}

\usepackage{booktabs, multirow} 
\usepackage{soul}
\usepackage[table]{xcolor} 

\usepackage[pagebackref=true,breaklinks=true,letterpaper=true,colorlinks,bookmarks=false]{hyperref}

\def\wacvPaperID{0642} 

\wacvfinalcopy 

\ifwacvfinal
\fi

\ifwacvfinal
\usepackage[breaklinks=true,bookmarks=false]{hyperref}
\else
\usepackage[pagebackref=false,breaklinks=true,colorlinks,bookmarks=false]{hyperref}
\fi

\begin{document}

\title{Dynamic Iterative Refinement for 
Efficient 3D Hand Pose Estimation}

\author{John Yang$^1$\thanks{Work done as an intern at Qualcomm Technologies, Inc.}, Yash Bhalgat$^2$, Simyung Chang$^3$, Fatih Porikli$^2$, Nojun Kwak$^1$\thanks{Corresponding Author}\\

\and
$^1$Seoul National University, Seoul, Korea \\
$^2$Qualcomm AI Research, Qualcomm Technologies, Inc., San Diego, CA, US\thanks{Qualcomm AI Research is an initiative of Qualcomm Technologies, Inc.}\\
$^3$Qualcomm AI Research, Qualcomm Korea YH, Seoul, Korea\\
{\tt\small \{yjohn, nojunk\}@snu.ac.kr, \{ybhalgat, simychan, fporikli\}@qti.qualcomm.com}
}
\maketitle
\thispagestyle{empty}

\begin{abstract}

While hand pose estimation is a critical component of most interactive extended reality and gesture recognition systems, contemporary approaches are not optimized for computational and memory efficiency. In this paper, we propose a tiny deep neural network of which partial layers are recursively exploited for refining its previous estimations. During its iterative refinements, we employ learned gating criteria to decide whether to exit from the weight-sharing loop, allowing per-sample adaptation in our model. Our network is trained to be aware of the uncertainty in its current predictions to efficiently gate at each iteration, estimating variances after each loop for its keypoint estimates. Additionally, we investigate the effectiveness of end-to-end and progressive training protocols for our recursive structure on maximizing the model capacity. With the proposed setting, our method consistently outperforms state-of-the-art 2D/3D hand pose estimation approaches in terms of both accuracy and efficiency for widely used benchmarks.
\end{abstract}


\section{Introduction}

Hand pose estimation (HPE) is an essential task for augmented reality and virtual reality (collectively called as “extended reality (XR)”) systems. For instance, to enable hand-based interactions with objects in XR environments, accurate real-time estimates of the positions of hand joints in 3D world coordinates are needed. Since hand gestures reflect elementary human behavioral patterns, hand pose tracking enables several downstream AI applications such as gesture recognition \cite{jang20153d,piumsomboon2013user,song2014air} and human-computer interactions \cite{chang2016spatio,jaimes2007multimodal}. Although many state-of-the-art HPE approaches \cite{cai2018weakly, cai2019exploiting, liu2019feature, tekin2019h+, yang2020seqhand, zimmermann2017learning} achieve high accuracy, they rely on large and complex model architectures, which incur a substantial computational cost. Therefore, such models are often unsuitable for relatively low-power computing systems, like wearables or hand-held XR devices \cite{arimatsu2020evaluation,du2020depthlab,portalware}.

To meet the constraints of resource-limited devices, \cite{huang2017multi,graves2016adaptive_act, veit2018convolutional, wang2018skipnet} have proposed neural network architectures with dynamic inference graphs that are conditional on the input. Recently, \cite{fan2020acenet} introduced an analogous approach for 3D HPE that exploits a Gaussian-kernel-based gating mechanism to adaptively combine the predictions of coarse and fine pose encoders, thus achieving a reduction in the GFLOPs for inference. While it alleviates the run-time complexity, the memory size required to deploy both pose encoders makes the method still infeasible for memory-constrained real-world applications. In this work, we tackle both the run-time efficiency and memory usage challenges of 3D HPE with a modular network whose capacity can be dynamically amplified through recursive exploitation of the network's parameters.

\begin{figure*}[t]
\centering
\includegraphics[width=0.89\textwidth, trim={.05cm 0.03cm .5cm 0.15cm},clip]{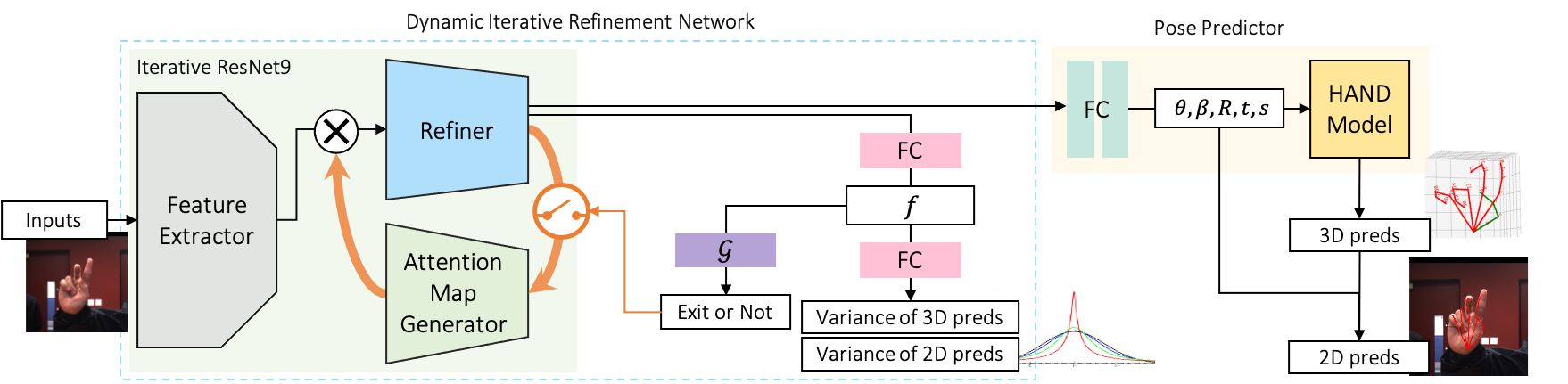}
\caption{\textbf{Dynamic Iterative Refinement Network (DIR-Net).} DIR-Net is composed of an \textit{Iterative ResNet9} (IR9) backbone network,
an \textit{Uncertainty Estimation module} and a \textit{Gating function}.
The backbone IR9 consists of a \textit{Feature Extractor} (FE), a \textit{Refiner} (RF) and an \textit{Attention Map Generator} (AMG).
For an image input, DIR-Net predicts 2D/3D hand joint locations. The model then refines its predictions with a repeated use of RF and AMG. Such re-use of RF $+$ AMG is continued until the gating function tells to exit the loop.
DIR-Net's inference is efficient in terms of both computation and memory due to its lightweight design. 
}
\label{fig:adapiterscopenet}
\vspace{-3mm}
\end{figure*}

Recursive usage of layers with shared parameters has been proposed by various works \cite{guo2019drnn, leroux2018iamnn, saha2020rnnpool, yoo2019extd} as a tool to match the performance of deeper networks with fewer parameters. Considering the distinctive semantic encoding at each layer in a neural network \cite{zeiler2014visualizing}, recursively exploited layers must manage gaps between higher-level features and lower-level ones. Thus, rather than directly re-feeding the output features to the consecutive recursive layers, we employ `refinements' of original input features using novel attention-augmentation (more details in Sec.\ \ref{sec:backbone}). This allows the recursive component of the deep network to focus on a distinct level of semantic information in each iteration.
 
With the above points in mind, we present Dynamic Iterative Refinement Network ({DIR-Net}), a modular weight-sharing network of which the capacity is adaptively amplified through dynamic recursive use of partial layers. DIR-Net is composed of a few components: \textit{Refiner} that handles distribution shifts in the attention-augmented input feature maps and produces refined predictions, \textit{Attention Map Generator} that refines the input features, and \textit{Gating Function} 
that dynamically controls the number of iterations. Along with the localized keypoint (joint) estimations, our backbone network is also trained to estimate the uncertainty in its own predictions. At every iteration, the gating function relies on these uncertainty estimations to decide whether the network should refine its current keypoint estimations or exit. This leads to per-sample adaptive inference in our model. 


Overall, our main contributions in this work are:
\begin{itemize}
\vspace{-2.5mm}
\item We introduce a lightweight architecture, DIR-Net, parts of which are utilized recursively while incorporating attention-augmentation and gating for dynamic refinement of the hand pose estimations.
\vspace{-2.5mm}
\item We propose a variance-based and also a reinforcement learning approach for dynamic gating that directly and indirectly exploits the uncertainty predictions of the estimated 2D/3D pose.
\vspace{-2.5mm}
\item We investigate the effectiveness of progressive and end-to-end training protocols for the inference-time efficiency of our recursive structure.
\vspace{-2.5mm}
\item With a comprehensive set of experiments and ablation studies, we show that our method achieves state-of-the-art performance in terms of both accuracy and efficiency on two widely used HPE benchmarks.
\end{itemize}

\section{Related Work}
\noindent\textbf{Adaptive Neural Computing.}
While going deeper in neural networks has helped achieve state-of-the-art performance on several image recognition tasks, early research works \cite{teerapittayanon2016branchynet,panda2016conditional} have pointed out that the task-specific complexity varies widely across input samples, and only a small fraction of inputs require processing by the entire network. For enabling energy-efficient implementations of deep networks, \cite{huang2017multi,liu2018dynamic, mullapudi2018hydranets} have proposed selecting different inference paths conditional on the input to optimize the overall trade-off between efficiency and accuracy. Residual networks \cite{resnet,he2016identity} have been a common choice as the backbone architecture for many dynamic inference approaches either through adaptive block removals \cite{wu2018blockdrop} or skips \cite{lee2020urnet, veit2018convolutional, wang2018skipnet}. Such dynamic gating turns on-and-off some of the blocks so that a network outputs its predictions through shortened forwarding paths. 

Similar to the idea of shortening the inference paths, early-exiting methods adaptively exit from neural networks before the inference reaches its final layer \cite{leroux2015resource-contrained, teerapittayanon2016branchynet}. For classification tasks, multiple classifiers are added along the feed-forward path. There is an early exit mechanism from the main inference path if the confidence of the earlier classifiers satisfies a pre-defined exiting criterion. These methods typically require heuristic values of confidence \cite{leroux2015resource-contrained} or entropy \cite{teerapittayanon2016branchynet}. Adaptive Neural Trees \cite{tanno2019adaptive_neuraltree} perform both dynamic gating and early exiting as its inference path is determined by the decisions at the leaves of a neural tree.


\noindent\textbf{Recurrent architectures for Image Recognition.}
Recurrent use of network layers with shared parameters has shown superior efficiency in a few recent works. Yoo \textit{et al.}~\cite{yoo2019extd} propose an iterative feature map generation method in which feature maps in different resolutions are iteratively generated by recurrently passing a network structured with inverted residual blocks. RNNPool \cite{saha2020rnnpool} also uses a recursive inference for efficient downsampling of the feature maps. Their method and ours share the notion of effectively bringing the higher-level feature information to the lower-level to overcome the limited semantic information in the lower-level features due to the shallow overall structure. However, their recurrent use of parameters does not work in an adaptive manner, operating with a fixed inference graph for all inputs.

As an early work, adaptive and iterative use of a network is proposed in \cite{graves2016adaptive_act} with the condition of a \textit{ponder} cost. The recurrent inference is adaptively repeated based on computation cost allowance. The later adaptive and iterative inference methods are proposed mainly with blocks of deep residual networks \cite{guo2019drnn, leroux2018iamnn}. While the original residual block is recursively exploited in \cite{guo2019drnn}, a neural recursive module that is inspired by residual blocks is introduced in \cite{leroux2018iamnn}. Our work is similar to the works of \cite{guo2019drnn, leroux2018iamnn} where parts of a model are recursively iterated to maximize the capacity of the parameters. 

\vspace{1mm}

\noindent\textbf{Adaptive 3D HPE.}
3D hand pose estimation problems have been actively studied during the last decade \cite{cai2018weakly, iqbal2018hand, mueller2018ganerated, yang2020seqhand, zimmermann2017learning}, including pre-deep-learning methods \cite{panteleris2018using, zhang20163d_pso_icppso}. However, despite its importance, 3D hand pose estimation methods have been rarely considered with adaptive computing. Recently, ACE-Net is proposed for efficient hand tracking \cite{fan2020acenet}. It is largely composed of two modular networks: one shallow, coarse pose encoder and another with a deeper, fine pose encoder. ACE-Net adaptively selects either encoder to feed-forward an input image. While this reduces computational cost during test time, the memory requirement for deploying a network with two separate encoders is infeasible for resource-constrained scenarios.

\section{Iterative Refinement Network}
In this section, we introduce our Iterative Refinement network, a modular weight-sharing neural model with iterative exploitation of parameters that yield refined 3D hand pose estimations in every iteration via attention-augmentation. As illustrated in Figure \ref{fig:adapiterscopenet}, our proposed overall architecture mainly consists of {Feature Extractor} (FE), {Refiner} (RF), {Attention Map Generator} (AMG), Gating Function ($\mathcal{G}$), an uncertainty estimation module and the 2D/3D hand pose predictor. During inference, a monocular RGB image of a hand is passed through the FE and sequentially to RF. The outputs of RF are then fed to (1) 2D/3D hand pose predictor, (2) uncertainty estimator, and (3) AMG (conditional on the gating function's output). 

Note that, in Sec.\ \ref{sec:exiting}, we describe two possible ways to perform the gating - (1) using a simple heuristic based on the uncertainty outputs, or (2) an RL agent ($\mathcal{G}$) trained using a reward function that optimizes the accuracy-efficiency trade-off. Only the latter case is illustrated in Figure \ref{fig:adapiterscopenet}.

\begin{figure}[t!]
    \begin{subfigure}{0.20\textwidth}
        \centering
        \includegraphics[height=4in]{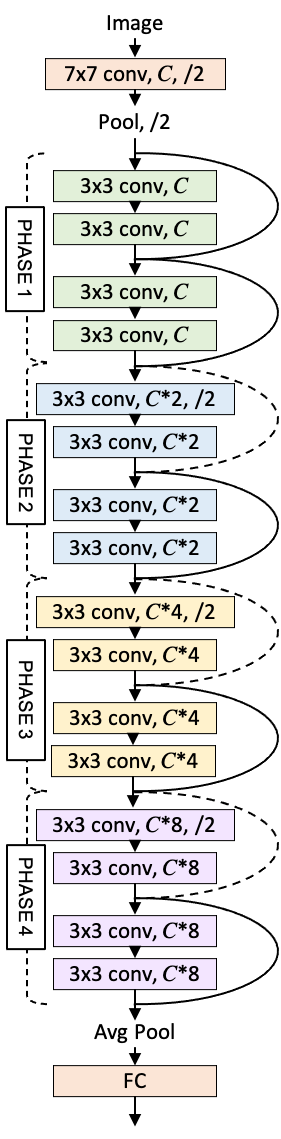}
        \vskip -1mm
        \caption{ResNet18}
    \end{subfigure}%
    ~ 
    \begin{subfigure}{0.24\textwidth}
    \centering
        \includegraphics[height=4in]{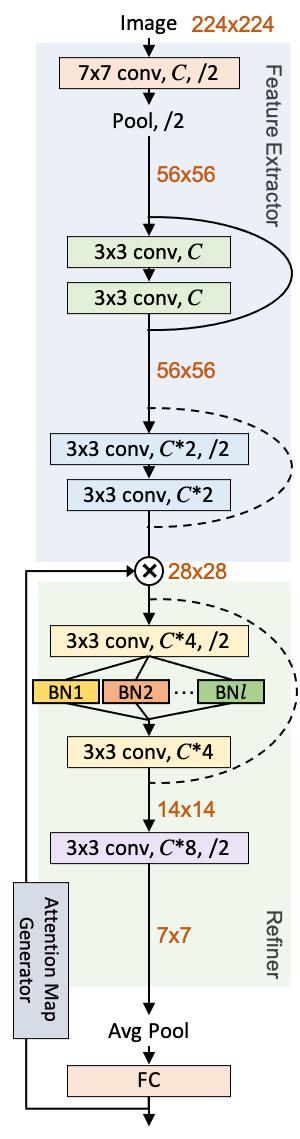}
        \vskip -1mm
        \caption{Iterative ResNet9}
    \end{subfigure}
    \vskip -2mm
    \caption{
    Iterative ResNet9 (IR9) follows a modular structure similar to ResNet18. The intention of the Refiner and Attention Map Generator is to reduce the overall size and complexity of ResNet18 while preserving the task-specific performance. IR9 uses distinct BatchNorm layers to handle distribution shifts in every iteration.
    }
    \label{fig:iter_resnet9}
    \vspace{-3mm}
\end{figure}

\subsection{Iterative ResNet9 Backbone} \label{sec:backbone}
Recursive inference allows for the usage of higher-level features to refine lower-level features in order to exploit the capacity of network parameters maximally. Such utilization of the network parameters enables a network with a much smaller size to have similar prediction accuracy as that of a more complex network \cite{guo2019drnn, leroux2018iamnn, saha2020rnnpool, zhang2019end}. Our tiny backbone network, where the default model capacity is low, is amplified with each iteration of recursive use. 

Inspired by ResNet18 \cite{resnet}, we design a modular network, ResNet9, with reasonably low complexity while keeping the downsampling characteristic of its original structure. Such modularization can be made with any network to exploit our proposed recursive refinement approach. We opted for a residual architecture due to its easily separable structure of block units, as done in relevant works \cite{guo2019drnn, lee2020urnet, leroux2018iamnn, yoo2019extd}.
As shown in Figure \ref{fig:iter_resnet9}b, the outputs of the ResNet9 backbone are refined by an iterative usage of the last few layers of the network. With such a mechanism, the proposed Iterative ResNet9 (IR9) requires much lesser memory to be deployed in systems due to its overall small model size. The components of this iterative backbone architecture are described in more detail below.

\noindent\textbf{Feature Extractor (FE):}
This component of the backbone network is responsible for encoding low-level image features. 
Since Feature Extractor is not operated recursively, its feature encoding involves heavy down-sampling of the feature maps to reduce the cost of the recursive computation that follows. In our best\footnote{In Sec.\ \ref{sec:ablation}, we provide a detailed ablation study of other downsampling ratios and their respective computation vs accuracy trade-offs.} model, FE reduces the spatial dimensions of image inputs from $224 \times 224$ to $28\times 28$.

\noindent\textbf{Refiner (RF):}
This part is recursively ``looped" in our proposed architecture. In every loop iteration (except the first), the attention-augmented feature maps are fed to this module, which refines the predictions of the previous iteration. Due to attention-augmentation, the distribution of the input feature map changes in each iteration. Hence, a separate Batch Normalization layer \cite{ioffe2015batch} is used in each iteration to handle the statistic shifts in input feature maps that are attention-augmented \cite{guo2019drnn, leroux2018iamnn}.

\begin{figure}[t]
    \centering
        \includegraphics[width=\linewidth]{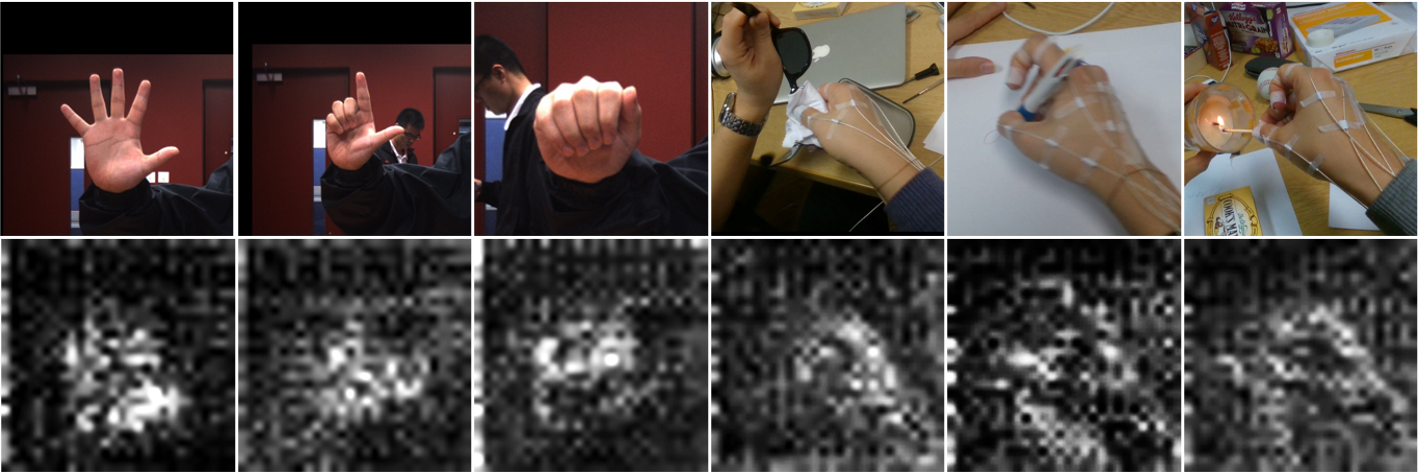}
    \vspace{-4.5mm}
    \caption{Images in the top row are the inputs and ones in the bottom row are normalized features that are computed with attention maps.}
    \label{fig:attmaps}
    \vspace{-4mm}
\end{figure}

\noindent\textbf{Attention Map Generator (AMG):}
This module has an upsampling decoder architecture that outputs an attention map tensor with values in the range $[0, 1]$ and of the same size as the feature map output of FE.
To effectively upscale the outputs of RF without any significant computational cost, the decoder is mainly composed of pixel-shuffle layers \cite{shi2016real}, which transfer the channel-wise information to the spatial domain with a negligible cost.
The main purpose of AMG is to merge information across different depths of the backbone network by spatially re-projecting the higher-level features to the lower-level feature maps. In doing so, the attention map tensor is element-wise multiplied with the output feature maps of FE. This allows dense connectivity that spatially links higher-level features and lower-level 2D feature planes \cite{huang2017multi, shi2016real}. Figure \ref{fig:attmaps} shows the effect of the attention maps on FE's output feature maps. The images in the bottom row shows the Hadamard product of FE feature output and attention map, averaged over channel axis and normalized.

\subsection{Pose Predictor} 
The structure of our pose predictor mainly follows the works of \cite{boukhayma20193d, yang2020seqhand}, consisting of two fully connected layers and a MANO hand mesh deformation model \cite{MANO}. MANO model takes low-dimensional pose and shape embeddings, respectively $\theta$ and $\beta$, as inputs for controlling the 3D hand mesh outputs: $J(\theta, \beta) = R_{\theta}(J(\beta))$.
The location of joints $J(\beta)$ can be globally rotated based on the pose $\theta$, denoted as $R_\theta$, to obtain a final hand model with corresponding 3D coordinates of 21 joints of a hand.

Our method takes cropped hand images as inputs $x$. The output of FE (denoted by $\mathcal{F}(x)$) is fed to RF $\mathcal{R}(.)$ along with an attention map $\mathcal{M}^l$ generated at each recursive iteration $l\in \{0, 1,2,...,l_{max}\}$. Our pose predictor takes $\mathcal{R}(\mathcal{F}(x))$ as input when $l=0$ and $\mathcal{R}(\mathcal{F}(x), \mathcal{M}^l)$ as input when $l>0$, and predicts a rotation matrix $R \in SO(3)$, a translation $t\in \mathbb{R}^2$ and a scaling factor $s \in \mathbb{R}^+$ along with its pose $\theta$ and shape $\beta$:
\begin{equation}
    \theta, \beta, R, t, s = \begin{cases}
            \Psi_{pose}(\mathcal{R}(\mathcal{F}(x))), & l=0\\
            \Psi_{pose}(\mathcal{R}(\mathcal{F}(x), \mathcal{M}^l)), & l>0,
          \end{cases}
\end{equation}
where $\Psi_{pose}(\cdot)$ represents a neural network with two fully-connected layers. 3D locations of joints $J(\theta, \beta)$ 
are obtained from MANO, and the 2D keypoint estimates are obtained by re-projecting these 3D locations to the 2D image plane with a weak-perspective camera model using the estimated camera parameters \{$R, t, s$\}:
\begin{equation}
    \begin{split}
        \hat{J}_{3D} &= J(\theta, \beta)\\
        \hat{J}_{2D} &= s\Pi R J(\theta, \beta) + t
    \end{split}
\end{equation}
where $\Pi$ represents orthographic projections.

With the re-projected 2D joint location estimations, the network can implicitly learn 3D joint locations with 2D labels \cite{boukhayma20193d, yang2020seqhand}. To train the pose predictor, we use $L1$ and $L2$ losses for 2D and 3D predictions respectively:
\begin{equation}
    \begin{split}
        L_{2D} &= ||  J_{2D}^{gt} - \hat{J}_{2D}||_1,\\
        L_{3D} &= ||  J_{3D}^{gt} - \hat{J}_{3D}||_2.
    \end{split}
    \label{eq:Pose_loss}
\end{equation}
Since the error for 2D estimations is calculated at pixel-level in image planes, $L_{2D}$ is desired to be more robust. The combination of using $L_1$-norm for $L_{2D}$ and $L_2$-norm for $L_{3D}$ has been found to be the best \cite{boukhayma20193d, girshick2015fastrcnn, he2019bbox}. We do not use the vertices of MANO hand mesh for training because their ground truths are not available during experiments. 
For the final loss, we also include objective terms that regularize pose and shape parameters, as done in \cite{boukhayma20193d}.

\section{Dynamic Exiting Mechanisms}
\label{sec:exiting}
Although a \textit{maximum} number of loop iterations is set in our Iterative Refinement network, not all input images require that many iterations of refinement. We propose alternative gating policies to determine when to stop iterating for each sample. The resulting architecture is called Dynamic Iterative Refinement Network (DIR-Net). Overall, two gating policies are proposed: (1) based on heuristic variance thresholds and (2) decisions of a reinforcement learning agent.

\subsection{Uncertainty Based Exiting} 
To decide whether to proceed to a next loop of recursive inference of the model, the model should be aware how certain its current predictions are. To this end, the model estimates variances for its 2D/3D keypoint predictions, by estimating a probability distribution instead of only joint locations. We simplify the problem and assume coordinates of joints are independent so we can use univariate Gaussians:
\begin{equation}
    P_W(J)= \frac{1}{\sqrt{2\pi\sigma^2}}e^{\frac{(J-\hat{J})^2}{2\sigma^2}}
\end{equation}
where $W$ refers to the trainable weights used to estimate $\sigma^2$. $J$ denotes a joint location coordinate and $\hat{J}$ represents the estimated joint location. The smaller the standard deviation $\sigma$ is, the more confident the model is with its own estimation. The ground truth coordinates are assumed to follow a Dirac-Delta distribution (i.e.\ $\sigma \xrightarrow{} 0$):
\begin{equation}
    P_D(J)= \delta(J-J^{gt}).
\end{equation}
Our model aims to minimize KL-Divergence between $P_W(J)$ and $P_D(J)$ to learn confidence estimation \cite{he2019bbox,seo2020kl}:
\begin{equation}
\begin{split}
    L_{conf} & = D_{KL}(P_D(J)||P_W(J))\\
            & \propto \frac{e^{-\alpha}}{2}(J^{gt} - \hat{J}) ^2 + \frac{1}{2}\alpha
    \end{split}
    \label{eq:conf_loss}
\end{equation}
where $\alpha\triangleq\log(\sigma^2)$. 

In practice, our network predicts the $\alpha$'s using a two-layer neural network $\Psi_{var}(\cdot) = \Psi_{var}^2(\Psi_{var}^1(\cdot))$ (shown as pink blocks in Figure \ref{fig:adapiterscopenet}):
\begin{equation}
    \alpha_{2D}, \alpha_{3D} = \begin{cases}
            \Psi_{var}(\mathcal{R}(\mathcal{F}(x))), & l=0\\
            \Psi_{var}(\mathcal{R}(\mathcal{F}(x), \mathcal{M}^l)), & l>0
          \end{cases}
    \label{eq:vari_output}
\end{equation}

\begin{figure}[t]
\centering
\includegraphics[width=0.97\linewidth, trim={.0cm 0.cm .0cm 0.cm},clip]{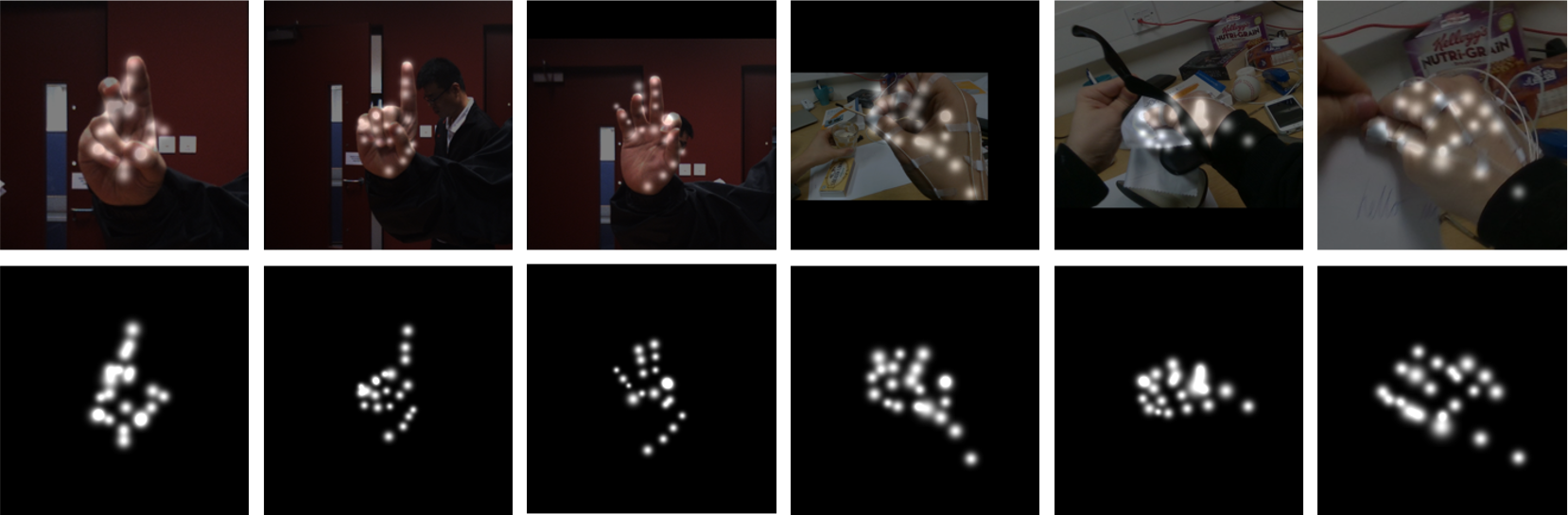}
\vspace{-2mm}
\caption{Heatmaps generated based on uncertainty for its 2D joint localization estimations. }
\label{fig:variances}
\vspace{-3mm}
\end{figure}

Following Eq. (\ref{eq:conf_loss}), the loss for variance estimation of 3D predictions is defined as: 
\begin{equation}
    L_{var_{3D}} = \frac{e^{-\alpha_{3D}}}{2}L_{3D}+ \frac{1}{2}\alpha_{3D}.
    \label{eq:var_3}
\end{equation}
Since we regress 2D joint locations with a smooth $L_1$ loss, we can define the loss for variance of 2D joint estimations as done in \cite{girshick2015fastrcnn, he2019bbox}: 
\begin{equation}
    L_{var_{2D}} = e^{-\alpha_{2D}}(L_{2D} - \frac{1}{2}) + \frac{1}{2}\alpha_{2D}.
    \label{eq:var_2}
\end{equation}

The total uncertainty loss ($L_{var}$) is defined as the sum of
$L_{var_{2D}}$ and $L_{var_{3D}}$. Using $L_{var}$, we allow the network to learn and estimate variances as a vector,
while conventional 3D hand pose estimators use decoders with deconvolutional layers to directly estimate Gaussian heatmaps for estimating the confidence scores \cite{baek2019pushing, boukhayma20193d,  cai2018weakly, iqbal2018hand, mueller2018ganerated}. Notably, the exponential terms in the objective function work as adaptive weights between $L_{2D}$ and $L_{3D}$ during training \cite{kendall2018multi, zhao2021domain}. 

The estimated variances can be directly utilized for decisions of exiting. We set a threshold value $\tau_{var}$ for the average variance for current joint estimations. If the average variance is larger than the threshold $\tau_{var}$, that means the keypoint estimations can be further refined, therefore another loop of RF + AMG is performed. If average variance is lower than $\tau_{var}$, we exit the loop. Figure \ref{fig:variances} shows examples of heatmaps generated by the estimated $\sigma^2$.



\begin{algorithm}[t]
 \caption{Progressive Training Protocol}
\footnotesize
 \textbf{Inputs:} maximum number of loops $l_{max}$, training data $S= \{s_i\} = \{( x, J_{2D}^{gt}, J_{3D}^{gt} )_i\}$ and learning rate $\texttt{Lr}$ \\
 \For{$l_{prog}= 0 $ to $ l_{max}$}{
 Initialize DIR-Net \texttt{DIR}$_{w,l_{prog}}$ \\with $l_{prog} + 1$ batch-norm layers\\
  \If{$l_{prog}>0$}{
  \texttt{DIR}$_{w,l_{prog}} \xleftarrow{} \texttt{DIR}_{w,l_{prog}-1}$\\
  Detach FE from training\\
  Reduce \texttt{Lr} by $\frac{1}{10}$ except AMG
  }
  \While{NOT stop criterion}{
  \For{$s_i \in S$}{
   \For{$l=0$ to $l_{prog}$ }{
    $\hat{J}_{2D}^l, \hat{J}_{3D}^l, \alpha_{2D}^l, \alpha_{3D}^l =\texttt{DIR}_{w,l}(x)$
    }
    Calculate $L_{total}$ using Eq. \ref{eq:loss_total} \\
    Update $w$ based on $L_{total}$ by $\texttt{Lr}$
    }
    }
 }
 \label{algo:prog-tr}
\end{algorithm}

\subsection{Decision Gating Function}
As an alternative to the heuristic uncertainty threshold value that decides whether to exit or continue, we propose a decision gating function for the network to learn its optimal decisions. This gating function, which is a two-layer neural network, is trained using a reward that optimizes the accuracy-efficiency trade-off. For an input $x$ and attention maps generated at $l$-th loop $\mathcal{M}^l$, the gating function outputs a \textit{stochastic} categorical decision of exiting.

The gating function $\mathcal{G}$ takes the feature vector $f=\Psi_{var}^1(\cdot)$ from the uncertainty estimation module as inputs.
As shown in Figure \ref{fig:adapiterscopenet} and Eq. \ref{eq:vari_output}, $f$ is a resultant vector created by the input $x$ and attention map generated at the loop $\mathcal{M}^l$ while also being the most determinant factor for variance estimations $\alpha$. The feature vectors thus consider such information for exit decisions. 
To this end, we train the gating function $\mathcal{G}(A_l|f_l)$ with on-policy vanilla policy gradient updates for two categorical actions $A_l \in$ \{EXIT, NOT EXIT\}. The gradient update is given by:
\begin{equation}
    \nabla_\textit{w} H(\textit{w}) = \mathbb{E}_\mathcal{G}\left[ r_l \nabla_\textit{w} \log \mathcal{G}_\textit{w}(A_l|\alpha_l)  \right],
\end{equation}
where $\textit{w}$ represents learnable parameters of the gating neural network, $r_l$ represents the immediate reward signal for the current loop, and $H(\cdot)$ is the total expected reward. 

We design the reward signal $r_l$ as a combination of the loss and computational cost (GFLOPs) taken by current iterations:
\begin{equation}
    r_l =  - \lambda (L_{2D}^l + L_{3D}^l) - l\cdot \text{GFLOPs}
    \label{eq:reward_signals}
\end{equation}
where $\lambda$ is a scale constant. With such rewards, the policy ultimately tries to minimize the error in the pose estimations while also minimizing the computational cost required. 

The gating network is trained separately, after the rest of the network including the uncertainty estimation component has been trained. While training the gating module, the remaining parts of DIR-Net are detached from training.

\begin{figure}[t]
\centering
\begin{subfigure}{1.0\linewidth}
  \centering
  \includegraphics[width=65mm, height=40mm, trim=2mm 1mm 9mm 12mm, clip=true]{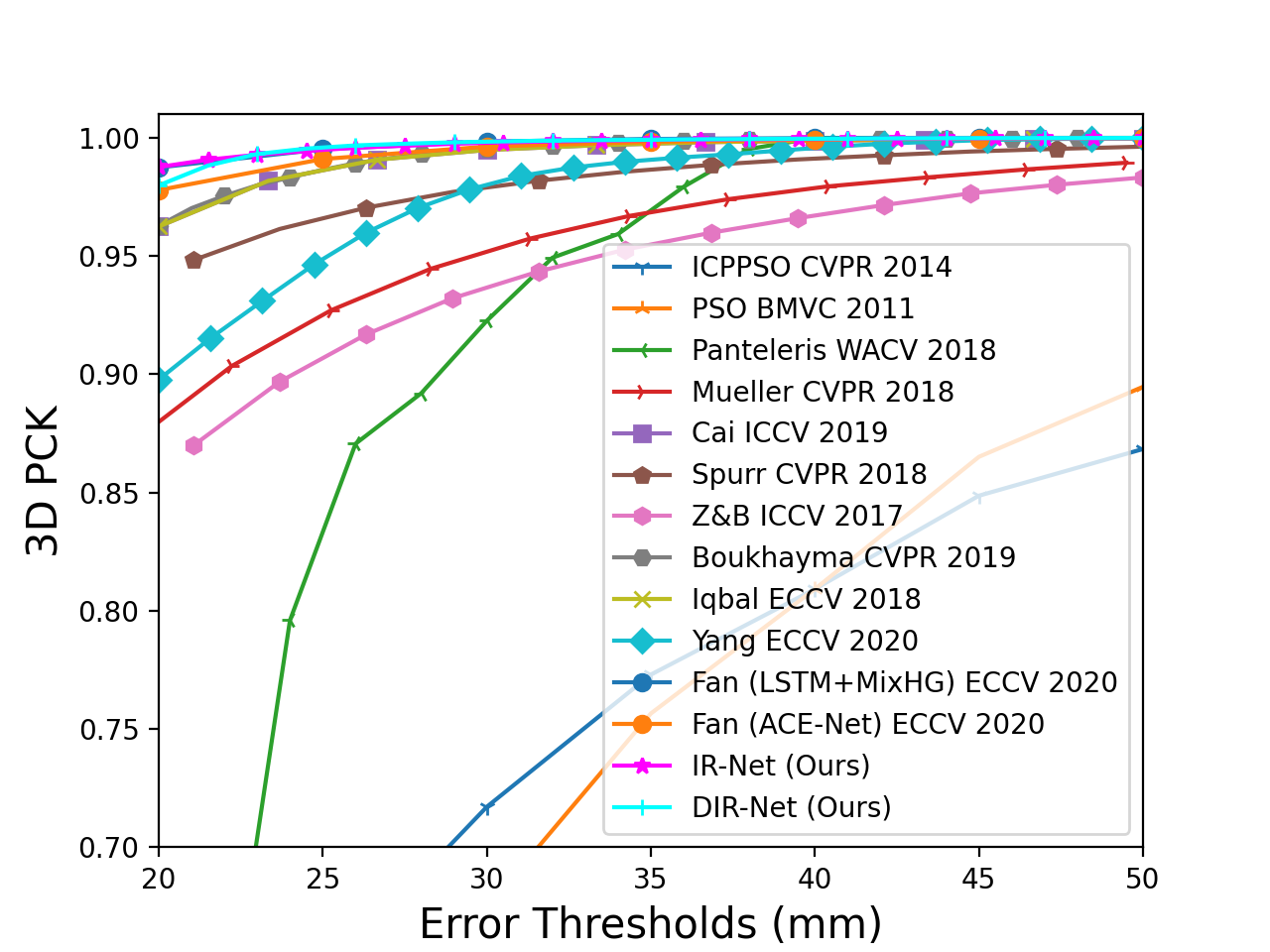}
  \caption{PCK for STB}
  \label{fig:pck_stb}
\end{subfigure}
\begin{subfigure}{1.0\linewidth}
  \centering
  \includegraphics[width=65mm, height=40mm, trim=2mm 1mm 9mm 12mm, clip=true]{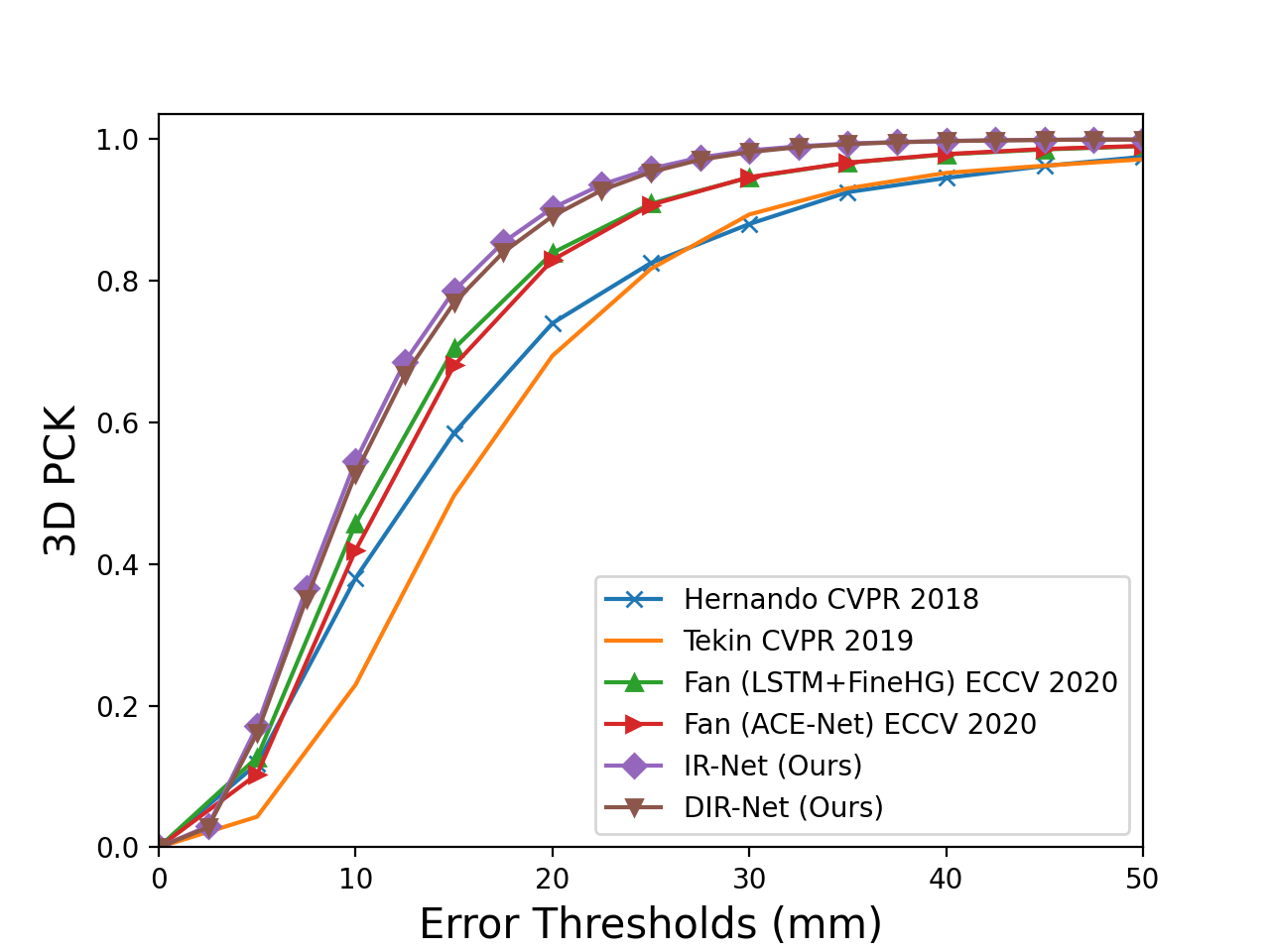}
  \caption{ PCK for FPHA}
  \label{fig:pck_fpha}
\end{subfigure}
\vspace{-2mm}
\caption{3D PCK Curve for STB and FPHA}
\vspace{-5mm}
\label{fig:pck}
\end{figure}

During inference, the gating function outputs a temperature-based softmax distribution \cite{jang2016gumbelsoftmax}, from which the actions are sampled. Using the softmax temperature parameter, $\tau_{gate}$, we can control the `harshness' of the exiting in our network. If $\tau_{gate}$ is higher, the model exits in the earlier loop iterations and therefore lesser number of FLOPs are used. As shown in Figure \ref{fig:taus}, by varying $\tau_{gate}$, we can obtain different accuracy vs computation trade-offs.
the parameter $\tau_{gate}$ allows expost-facto control over our model after training, which can be used to adjust the generated
action probabilities to change the number of loops after training the model.

\begin{table}[t]
\centering
\caption{SOTA Efficiency Comparison}
\vspace{-3mm}
\resizebox{0.81\linewidth}{!}{
\begin{tabular}{lccr}
\multicolumn{4}{c}{\textbf{STB}} \\
Methods &AUC(20-50)&GFLOPs &\#Params \\\toprule
Z\&B\cite{zimmermann2017learning} &0.948 &78.2 &\multicolumn{1}{c}{-} \\
Liu et al.\cite{liu2019feature} &0.964 &16 &\multicolumn{1}{c}{-} \\
HAMR\cite{zhang2019end} &0.982 &8 &\multicolumn{1}{c}{-} \\
Cai et al.\cite{cai2019exploiting}&0.995 &6.2 &4.53M \\
Fan et al.\cite{fan2020acenet} &0.996 &1.6 &$>$4.76M \\
\textbf{Ours }&\textbf{0.997} &1.31 &1.68M \\
\midrule
\multicolumn{4}{c}{\textbf{FPHA}} \\
Methods &AUC(0-50)&GFLOPs &\#Params \\\toprule
Tekin et al.\cite{tekin2019h+} &0.653 &13.62 &14.31M \\
Fan et al. \cite{fan2020acenet} &0.731 &1.37 &$>$5.76M \\
\textbf{Ours }&\textbf{0.768} &0.28 &460K \\
\bottomrule
\end{tabular}}\\
\vspace{-2mm}
\label{tab:SOTA}
\end{table}

\section{Progressive Training}
\label{sect:prog_tr}
In the training of our network, we try to minimize the sum of losses from all loops with a predefined maximum number of loops $l_{max}$ \cite{shi2016real}:
\begin{equation}
    L_{total} = \sum_{l=0}^{l_{max}} \gamma_{2D}L_{2D}^l +\gamma_{3D}L_{3D}^l + \gamma_{var}L_{var}^l.
    \label{eq:loss_total}
\end{equation}
This can be done in either an end-to-end or progressive manner. The end-to-end protocol trains once with a predefined maximum number of loops. The progressive protocol trains the network multiple times while incrementing the total number of loops progressively.

The progressive training protocol is summarized in Algorithm \ref{algo:prog-tr}. We train the network $l_{max}+1$ times, each time with the maximum number of loops $l_{prog}\in\{0,1,2,...,l_{max}\}$. The network is initially trained without any loop for the case of $l_{prog}=0$. This initial training phase requires one BatchNorm layer at the beginning of RF for a single inference path. Then, for every $l_{prog}>0$, DIR-Net is initialized with $l_{prog}+1$ number of BatchNorm layers, and the parameters that were learned in the $l_{prog}-1$ iteration are loaded into the network (except for the extra BatchNorm layer). Since the FE component of the network learns meaningful feature encoding layers when trained with $l_{prog}=0$, FE is detached from further learning when $l_{prog}>0$. The learning rate is reduced by a factor of $1/10$ in every iteration for the components of the network, except the AMG which is trained with the original base learning rate. The progressive training protocol is empirically shown to ensure maximization of network's capacity at each loop, yielding a higher frequency of exits at early loops hence lowering average computational cost for inference.

\section{Experiments}
\label{sec:experiments}


\noindent\textbf{Datasets.}\footnote{Datasets are acquired via requests with a university domain.}
STB dataset has real hand images sequentially captured in 18,000 frames with 6 different lighting conditions and backgrounds. Each frame image is labeled with 3D annotations of 21 joints. 
Along with the training set of STB, our model is trained with PANOPTIC datasets \cite{simon2017panoptic} and evaluated on the testing set of STB, as done in \cite{boukhayma20193d}. PANOPTIC dataset is re-engineered from data from multiple views of Panoptic studio \cite{joo2015panoptic}. The dataset is made of 14,847 image samples along with 2D joint annotations, and provides general views of hands and skin tones. 

FPHA dataset \cite{garcia2018fpha} consists of RGB video sequences of 6 subjects performing 45 types of hand activities with daily objects in egocentric views (e.g. pouring a bottle, charging a phone) that follow with heavy (self-)occlusions. Each sample is annotated with 2D and 3D labels both of which are used for training. We follow the official split of the dataset.

\vspace{1mm}
\noindent\textbf{Metrics.} For evaluation results, we measure the percentage of correct keypoints (PCK) for estimated 2D/3D joint locations and the area under the curve (AUC) of various error thresholds. In addition, we provide average Euclidean distance error for all 2D/3D joints for absolute comparisons.

\vspace{1mm}
\noindent\textbf{Implementation Details.} 
Following our main comparison \cite{fan2020acenet}, we also present two models that are structured differently for the datasets.
Since our work not only improves computational efficiency but also decreases the overall model size, we believe it is more objective to see our models in their smallest size possible that
follow with competitive performance.
Within IR9, 32 and 16 base channels (i.e. \# output channels in first layer) are used respectively for STB and FPHA dataset. For the fully connected layers, the number of nodes for each layer is 512 for STB and 256 for FPHA.
The initial learning rate is 10$^{-3}$ for both progressive and end-to-end trainings.
For the progressive protocol, the network is initially trained with neither looping nor use of AMG for 50 epochs. 
For loops $l>0$, the network is reset with new learning rates as described in Sec. \ref{sect:prog_tr} and trained for 20 epochs for each $l_{prog}$ case.
For end-to-end training protocol, the network is trained for the same amount of epochs for equivalent maximum loop training setting (e.g. $90=50+20+20$ epochs for $l_{max}=2$).

\begin{figure}[t]
    \centering
    \begin{subfigure}{\linewidth}
      \includegraphics[width=\linewidth]{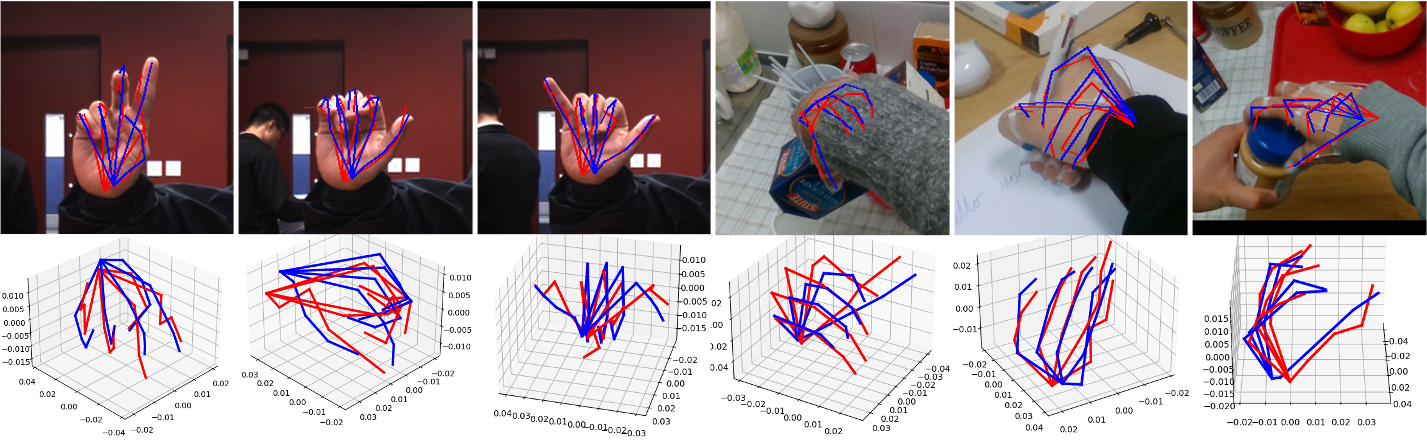}
    \end{subfigure}\\
    \begin{subfigure}{\linewidth}
        \centering
      \includegraphics[width=0.7\linewidth]{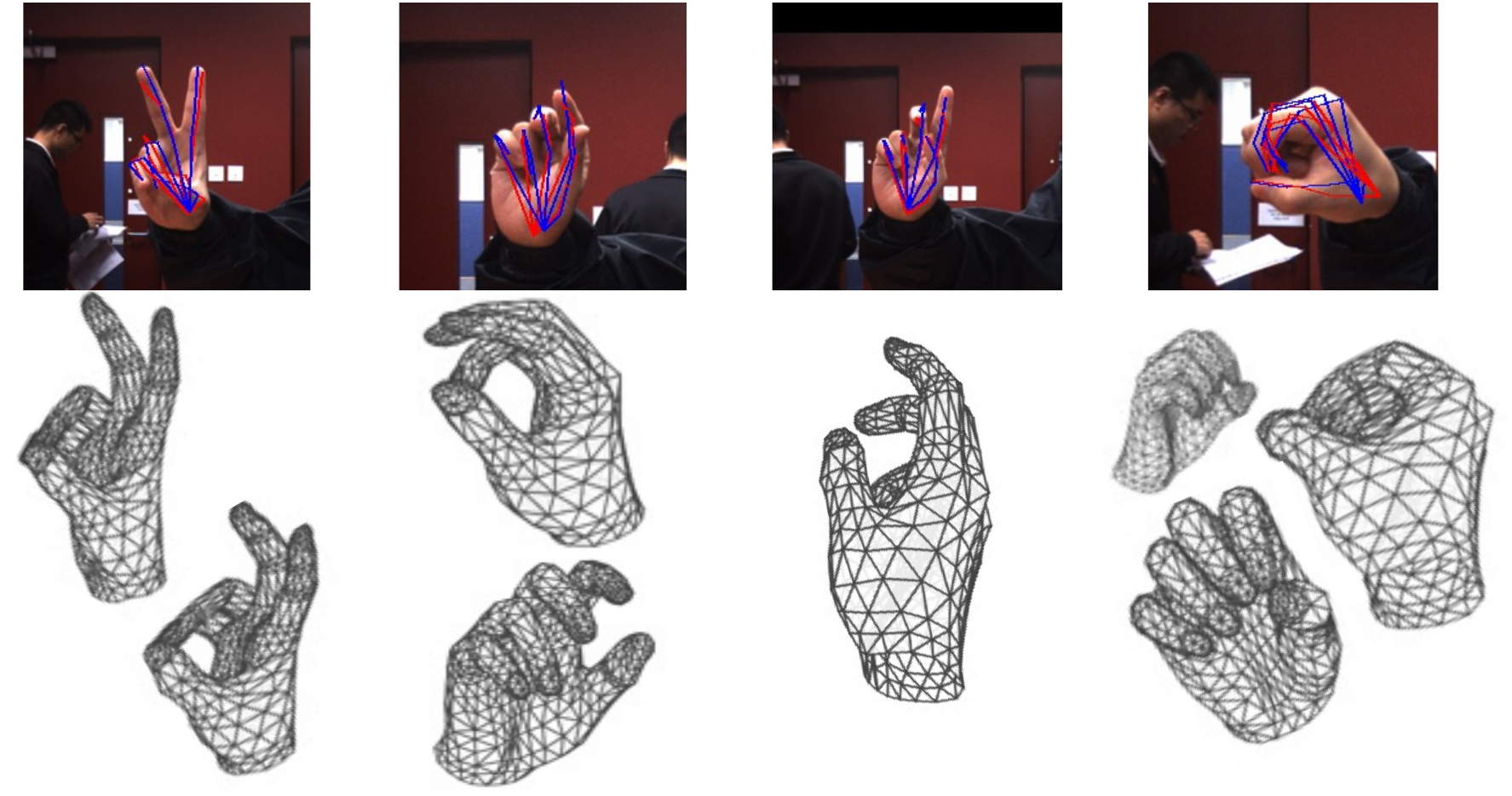}
    \end{subfigure}
    \vspace{-2mm}
    \caption{Qualitative results for STB and FPHA datasets}
    \label{fig:qual_res}
    \vspace{-2.5mm}
\end{figure}

\subsection{Comparison Against SOTA Methods}
Our methods without/with dynamic gating are respectively labeled as `IR-Net' and `DIR-Net' for comparison against relevant contemporary methods in Figure \ref{fig:pck}.
For STB dataset, entries include deep-learning based works of \cite{boukhayma20193d, cai2018weakly, fan2020acenet, mueller2018ganerated, spurr2018cross, yang2020seqhand, zhang2019end, zimmermann2017learning} and approaches from \cite{panteleris2018using, zhang20163d_pso_icppso}. 
Figure \ref{fig:pck_stb} shows that our method without adaptive gating performs the best of the entries with 3D AUC of 0.998. 
With adaptive gating, our method performs with AUC of 0.997 which outperforms the recent adaptive 3D HPE method \cite{fan2020acenet}.

For FPHA, our method is compared against contemporary methods including \cite{fan2020acenet, garcia2018fpha, tekin2019h+}, and outperforms them as plotted in Figure \ref{fig:pck_fpha}.
We believe large performance enhancement comes from our effective architecture with residual structure and kinematic fitting of predefined hand model, especially for FPHA dataset, yet our method allows to preserve the accuracy though of large reduction in model size and complexity. 
Qualitative results for both datasets are depicted in Figure \ref{fig:qual_res} and the Supplementary.

Our model's overall performance is compared to SOTA methods in Table \ref{tab:SOTA}. The work of \textit{Fan et al.} \cite{fan2020acenet} is a recent efficient method with attempts of reducing the model complexity for 3D HPE tasks, the size of which is reported partially instead of their whole model. Their coarse and fine pose encoders add up to what is reported in the table, which makes their overall model size even larger. The average of loops required for our method during validation is 4.2 for STB and 2.25 for FPHA.

\begin{figure}[t]
  \centering
  \includegraphics[width=\linewidth, trim={1.15cm 0cm 2.1cm 1.2cm},clip]{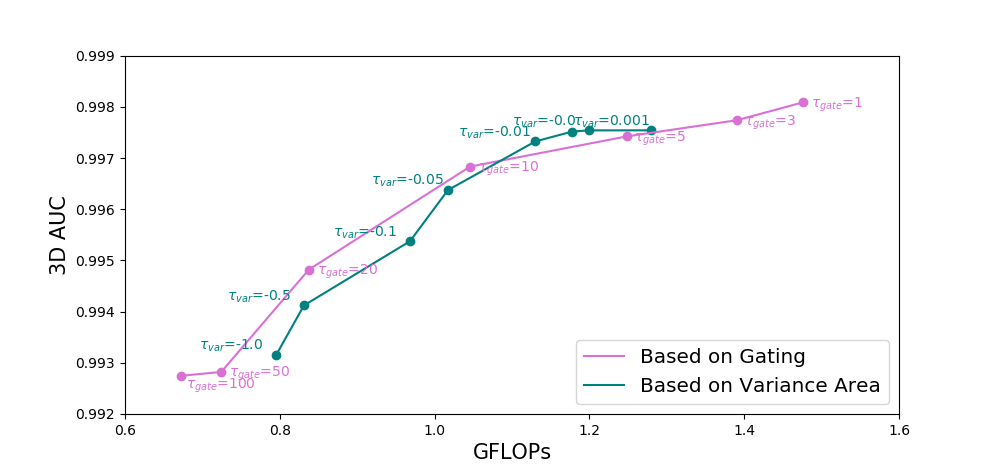}
  \vspace{-7.5mm}
  \captionof{figure}{Efficiency and accuracy trade-off for STB with various variance thresholds $\tau_{var}$ and temperature parameters $\tau_{gate}$.}
  \vspace{-4mm}
  \label{fig:taus}
\end{figure}

\vspace{1mm}
\noindent\textbf{Computational Efficiency.}
In Figure \ref{fig:taus}, we show the efficiency and accuracy trade-off of our iterative refinement model trained for STB dataset. Plots of 3D AUC for for various $\tau_{gate}$ represent the temperature parameter of our softmax policy gating outputs.
Higher values of $\tau_{gate}$ cause softer distribution of softmax.
Various heuristic values of $\tau_{var}$ are explored for the performance trade-offs.
Our gating function reaches higher overall performance than that of heuristic thresholds.
Meticulous control over $\tau_{var}$ is needed to reach competitive performances gained with $\tau_{gate}$ values.

\begin{figure}[t]
\centering
\begin{subfigure}{0.50\linewidth}
  \centering
  \includegraphics[height=5.3cm, trim=1mm 2mm 11mm 15mm, clip=true]{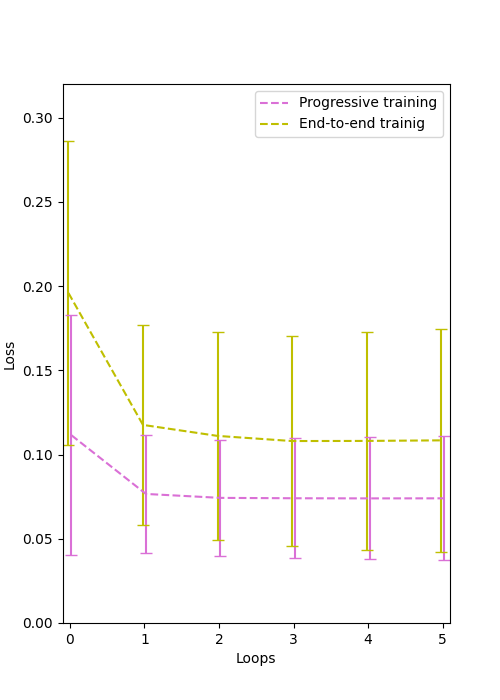}
  \vspace{-3mm}
  \label{fig:sub1}
    \caption{STB}
\end{subfigure}
\begin{subfigure}{0.43\linewidth}
  \centering
  \includegraphics[height=5.3cm, trim=6mm 2mm 11mm 15mm, clip=true]{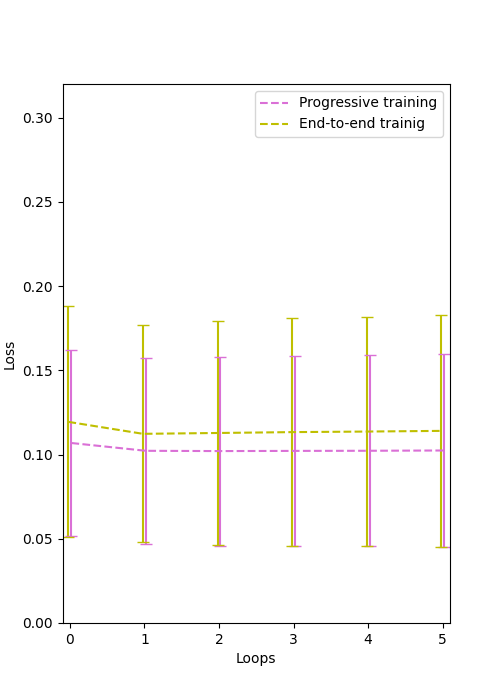}
  \label{fig:sub2}
     \vspace{-7mm}
    \caption{FPHA}
\end{subfigure}
\vspace{-3mm}
\caption{Range of losses of validation samples after various different numbers of recursive inferences.}
\label{fig:losses_over_loops}
\end{figure}

\begin{figure}[t]
\centering
\includegraphics[width=0.8\linewidth, trim={.0cm 0.cm .0cm 0.3cm},clip]{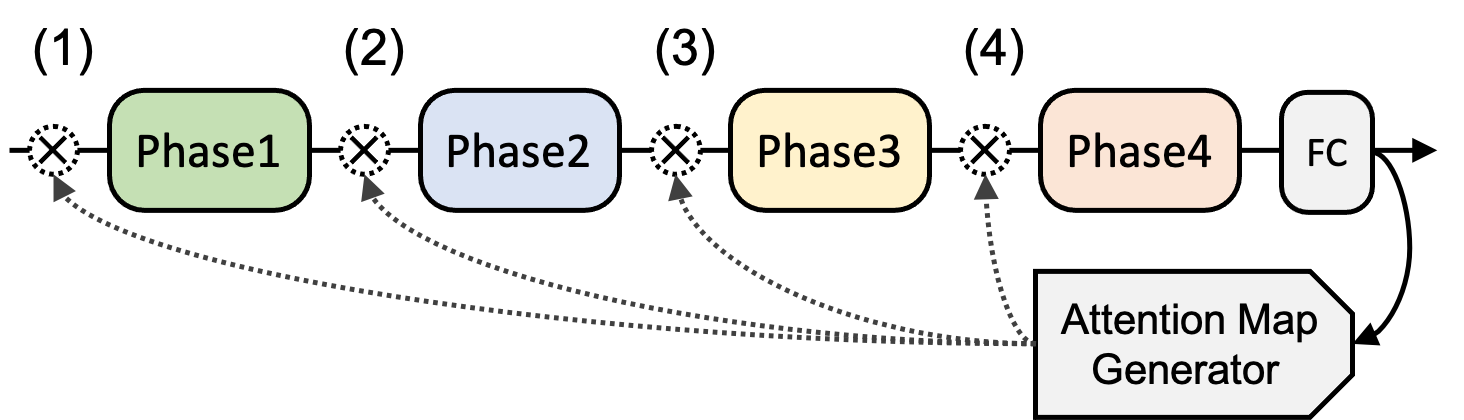}
\vspace{-1.5mm}
\caption{\small Possible loop points within IR9 structure, which also affects the structure/complexity of FE, RF and AMG.}
\vspace{-2.5mm}
\label{fig:loop_points}
\end{figure}

\subsection{Ablation Studies}
\label{sec:ablation}
For the recursive path of IR9, four possible points can be selected to perform iterative inference as depicted in Figure \ref{fig:loop_points}. 
Since output feature maps vary in sizes among the points, the structure of AMG also changes to generate corresponding spatial maps, which yields different computation cost for each loop point.
We provide pose estimation results of our proposed structure with various loop points in Table \ref{tab:loop_points}. 
Here, the results after the final recursive inference of DIR-Net are reported. Although we got the best 3D AUC with the second loop point for FPHA dataset, competitive 2D/3D estimations are achieved with the third loop point with lesser computation, hence providing the best trade-off.

The iterative refinement is evaluated with different structures in Table \ref{tab:ablation}. 
The baseline is set with our network without any iteration but just forwarding inference.
For iterations without AMG, 
the 7$\times$7 output features after 4th phase of IR9 (see Figure \ref{fig:iter_resnet9}b) are spatially enlarged with pixel-shuffle \cite{shi2016real} and up-sampling.
This method represents direct recursive use of higher-level features as done in \cite{guo2019drnn, leroux2018iamnn, saha2020rnnpool}. 
Performances of our network are reported with and without dynamic exits. 
Results show that recursive refinements effectively fuse higher-level features with lower-level ones for the overall best performance. 

Figure \ref{fig:losses_over_loops} presents the distribution of losses computed at each iteration, implicitly showing the performance difference made by our model trained in  progressive and end-to-end manners. 
End-to-end training of recursive structure carry the conventional gradient vanish problem \cite{bengio1994learning}, as also mentioned in a relevant work \cite{guo2019drnn}.
Such implication is provided with more comparison results of the progressive and end-to-end training protocols in the Supplementary.

\setlength{\tabcolsep}{3pt}

\begin{table}[t]
\centering

\caption{\small \textbf{Ablation studies on looping points} Prediction results based on 5th recursive inference of DIR-Net trained with $l_{max}=5$ are reported.}
\label{tab:loop_points}

\scriptsize
\centering
\vspace{-3mm}
\resizebox{0.655\linewidth}{!}{
\begin{tabular}{cccccc}
&\multicolumn{4}{c}{\textbf{STB}}&  \\ \toprule
Loop &\multicolumn{2}{|c}{AUC (20-50)}  &\multicolumn{2}{c|}{Err (px/mm)} & \multirow{2}{*}{GFLOPs}   \\
Points&\multicolumn{1}{|c}{2D}  &3D &2D &\multicolumn{1}{c|}{3D} &\\\midrule
(1) &\multicolumn{1}{|c}{0.751} &0.993 &7.81 &\multicolumn{1}{c|}{8.26} &  2.22 \\
(2) &\multicolumn{1}{|c}{0.791} &0.994 &6.12 &\multicolumn{1}{c|}{7.86} & 1.63 \\
(3) &\multicolumn{1}{|c}{\textbf{0.806}} &\textbf{0.998} &\textbf{5.93} &\multicolumn{1}{c|}{\textbf{6.88}} & 1.51\\
(4) &\multicolumn{1}{|c}{0.780} &0.996 &6.69 &\multicolumn{1}{c|}{7.39} & 0.86 \\
\midrule
\end{tabular}}

\centering
\resizebox{0.655\linewidth}{!}{
\begin{tabular}{cccccccccccc}
 &\multicolumn{4}{c}{\textbf{FPHA}} & \\ \toprule
Loop & \multicolumn{2}{|c}{AUC (0-50)} & \multicolumn{2}{c|}{Err (px/mm)}   & \multirow{2}{*}{GFLOPs}  \\
Points&\multicolumn{1}{|c}{2D}  &3D &2D &\multicolumn{1}{c|}{3D} \\\midrule
(1) &\multicolumn{1}{|c}{0.716}  &0.764 &8.68 & \multicolumn{1}{c|}{11.79} & 0.6\\
(2) &\multicolumn{1}{|c}{0.713}  &\textbf{0.775} &8.77 & \multicolumn{1}{c|}{\textbf{11.26}} & 0.5\\
(3) &\multicolumn{1}{|c}{\textbf{0.717}}  &0.772 & \textbf{8.64} & \multicolumn{1}{c|}{11.54} &0.45\\
(4) &\multicolumn{1}{|c}{0.716}  &0.767  &8.65 &  \multicolumn{1}{c|}{11.68} & 0.29\\
\bottomrule
\end{tabular}}
\vspace{-2.5mm}
\end{table}

\begin{table}[t]
\centering
\caption{\small \textbf{Ablation studies on recursive refinement structure}}
\vspace{-2mm}
\label{tab:ablation}
\resizebox{1.0\linewidth}{!}{
\begin{tabular}{lccccccc}
&\multicolumn{3}{c}{\textbf{STB}} &\multicolumn{3}{c}{\textbf{FPHA}} \\\cmidrule{2-7}
&3D AUC &GFLOPs &\#Params &3D AUC &GFLOPs &\#Params \\\midrule
No Iter. &0.987 &0.46 &1.44M &0.768 &0.14 &408K \\
Iter. w/out AMG &0.991 &1.06 &1.44M &0.769 &0.30 &408K \\
Iter. w/ AMG  &0.998 &1.51 &1.68M &0.772 &0.45 &460K \\
Iter. w/ AMG + $\mathcal{G}$  &0.997 &1.31 &1.68M &0.768 &0.28 &460K \\
\bottomrule
\end{tabular}}
\vspace{-3mm}
\end{table}

\section{Conclusion}
In this work, we propose DIR-Net, a lightweight network parts of which are utilized recursively for prediction refinements with adaptive scoping and dynamic gatings.
For gating criteria, we introduce an objective function that allows our method to estimate uncertainty of its own 2D/3D pose estimations. 
This allows dynamic exits for computation efficiency based on pre-defined heuristic thresholds for variance or the decision of the gating function.
We also investigate the effectiveness of progressive and end-to-end training protocols for our recursive structure.
Training a recursive network in a progressive manner with an increasing number of loop allowance is empirically shown to perform better than training in an end-to-end manner, maximizing capacity of parameters that are recursively exploited. 
The proposed method reaches the SOTA performance in terms of both accuracy and efficiency.

\vspace{3mm}
\noindent\textbf{Acknowledgement}
This work  was  supported  by  the  National  Research  Foundation of Korea (NRF) grant funded by the Korea government (2021R1A2C3006659).

{\small
\bibliographystyle{ieee_fullname}
\bibliography{egbib}
}

\end{document}